\documentclass[11pt]{article}
\usepackage[preprint]{acl}
\usepackage{times}
\usepackage{latexsym}
\usepackage[T1]{fontenc}
\usepackage[utf8]{inputenc}
\usepackage{microtype}
\usepackage{inconsolata}
\usepackage{graphicx}
\usepackage{booktabs}
\usepackage{amsmath}
\usepackage{fix-cm}
\usepackage{tabularx}
\usepackage{xcolor}
\usepackage[percent]{overpic}
\usepackage{fvextra}
\DeclareUnicodeCharacter{2192}{\ensuremath{\rightarrow}}
\newsavebox{\setupSkillBox}
\definecolor{HarnessWarm}{HTML}{FBF4E7}
\definecolor{HarnessSoft}{HTML}{F6F4EF}
\definecolor{HarnessBand}{HTML}{FAF9F6}
\definecolor{HarnessScore}{HTML}{EEF5F2}
\definecolor{HarnessDeny}{HTML}{FFF7F5}
\newcommand{\harnessFigurePatch}[7]{%
\put(#1,#2){\color{#5}\rule{#3\unitlength}{#4\unitlength}}%
\put(#1,#2){\raisebox{0.3\unitlength}{\makebox[#3\unitlength][l]{\color{black}\sffamily\fontsize{#6}{6pt}\selectfont #7}}}%
}

\title{Design and Cross-Domain Evaluation\\of an Agent Execution Harness}
\author{Saurav Joshi \quad Manas Gupta \quad Ashwin Venkatesh Raman \quad Akshay Budhkar\\
\normalfont Spine AI}

\begin{document}
\maketitle
\begin{abstract}
Completing complex tasks with language-model agents requires coordinating planning, execution, and review so the final result addresses the user's original objective. We present an agent execution harness that keeps the original objective separate from a revisable plan and an execution record. A planner assigns scoped tasks, workers produce outputs and artifacts, and a reviewer examines results against the original request before submitting or requesting additional work. Reusable domain-specific instructions guide these roles and are refined during construction using task inputs and execution trajectories, without official grading feedback. We evaluate complete harness configurations across terminal tasks, clinical responses, application building, and scientific questions. On Terminal-Bench 2.1, the harness passes 83 of 89 instances (93.26\%); on HealthBench Professional, a Fable-configured system achieves a length-adjusted rubric score of 69.63 across 525 tasks. In a 53-task paired HealthBench comparison, the Fable-configured harness achieves an aggregate score 19.89 points higher than direct completion, despite a $-1.46$-point median task-level change; mean inference cost rises from \$0.29 to \$2.13 per task. We combine these outcomes with latency, documented instruction revisions, and reviewer decisions. Evaluating outcomes alongside execution trajectories and resource use clarifies how coordination supports task completion, where it introduces regressions, and at what cost.
\end{abstract}

\section{Introduction}
\label{sec:introduction}

An agent can complete an assigned step without fulfilling the user's original objective. For example, a database query may return the expected rows with incorrect column types, or an application may appear functional until a reload exposes an unmet authentication requirement. Such cases illustrate the distinction between local execution and task completion. As tasks move through planning, execution, and review, a system must preserve the original request and the evidence needed to decide whether a result should be submitted or work should continue. Coordination and verification failures make this an important challenge for agent systems \citep{cemri2025mast}.

Tool-using agents support iterative reasoning and action \citep{yao2023react,yang2024sweagent}, while multi-agent systems provide role assignment, orchestration, and artifact handoff \citep{hong2024metagpt,qian2024chatdev,fourney2024magenticone}. We study how these capabilities are integrated into an execution protocol. A plan can change as new work is identified, whereas the user's original objective must remain the reference for completion. A worker's reported success must therefore be assessed in relation to the overall result.

We present an agent execution harness that coordinates planning, worker execution, and review through three elements: the original objective, a revisable graph of scoped assignments, and an accumulating execution record. A planner assigns work and dependencies; workers execute those assignments and record outputs and artifacts; and a reviewer examines the available results against the original objective before submitting or requesting bounded follow-up work. Benchmark-specific adapters connect this protocol to different tools, environments, and submission requirements.

Reusable domain-specific instructions guide the planning, execution, and review roles. During construction, a setup agent refines these instructions using task inputs and execution trajectories without official reference answers or grading feedback. The resulting instructions specialize the common execution protocol without changing its underlying structure.

We evaluate complete harness configurations across terminal tasks, clinical responses, application building, and scientific questions using benchmark-native metrics. We supplement these outcomes with fixed-task reference comparisons, cost and latency measurements where available, and analyses of documented instruction changes and reviewer behavior. Paired clinical comparisons show substantial improvements alongside regressions and increased resource use. Retrospective terminal episodes show review identifying errors and directing focused repairs, while a contrasting virtual-machine episode shows additional verification disrupting a working environment. These analyses connect benchmark outcomes to observed execution behavior and resource costs \citep{ma2024agentboard}.

We investigate three research questions:
\begin{enumerate}
    \item \textbf{Cross-domain outcomes and resources:} What task outcomes do complete harness configurations achieve across different domains, and what model expenditure and latency accompany them where measured?
    \item \textbf{Task-level comparisons:} How do harness configurations compare with reference configurations on fixed task subsets, and what improvements, regressions, and unchanged outcomes emerge in paired clinical comparisons?
    \item \textbf{Execution behavior:} What do recorded execution trajectories, reviewer decisions, and instruction revisions reveal about how the harness handles incomplete work, directs corrections, and encounters failures?
\end{enumerate}

This paper makes three contributions to the design, domain specialization, and empirical evaluation of agent execution systems:
\begin{enumerate}
    \item \textbf{An agent execution architecture} that specifies interfaces linking the original objective, revisable work graph, and artifact record to dependency-aware execution, benchmark-specific submission, and bounded reviewer-directed continuation.
    \item \textbf{A cross-domain empirical evaluation} of complete harness configurations across four task domains, combining benchmark-native outcomes, fixed-task comparisons, and available model cost and latency measurements.
    \item \textbf{An analysis of execution behavior} using documented instruction revisions, reviewer decisions, and task-level successes and failures to examine how the execution protocol operates in practice, including effective corrections and unsuccessful verification.
\end{enumerate}

\section{Method}
\label{sec:method}

The harness coordinates scoped planning, artifact-producing workers, and review against the original objective. Domain-specific instructions specialize these roles within a shared execution protocol.

\subsection{Goal, work graph, and execution record}

A mission consists of the original objective $\mathcal C$, a work graph $\mathcal G_t=(V_t,A_t)$, and an execution record $S_t$ at round $t$. The objective consists of the user's task request and benchmark instructions and serves as the reference for completion. The graph describes proposed work; the record contains task states, worker outputs, and artifact references. Keeping the objective separate from the revisable graph allows subsequent planning and review to return to the original request.

\begin{figure*}[t]
\centering
\begin{overpic}[width=\textwidth]{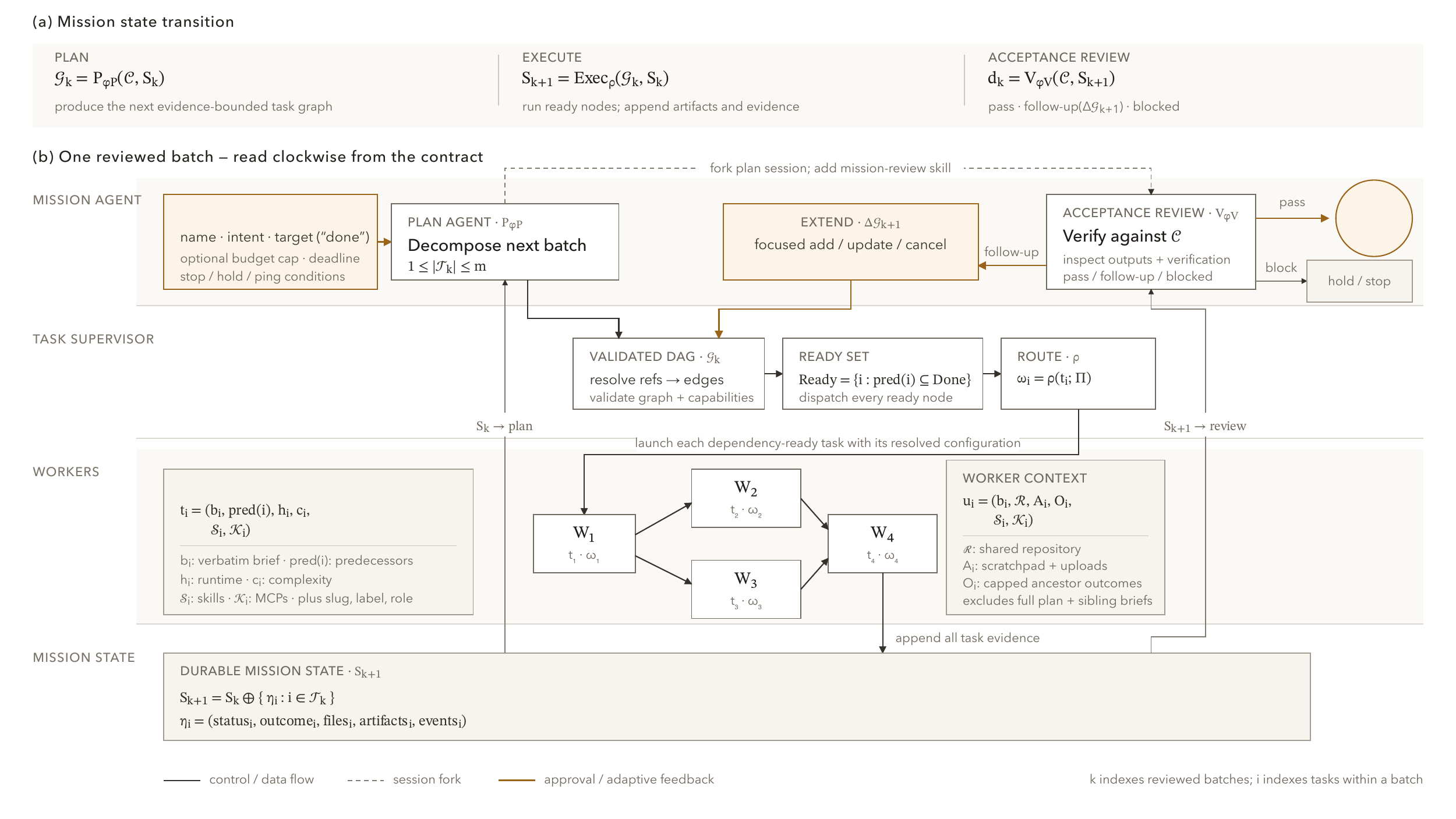}
\harnessFigurePatch{12.1}{40.6}{13.8}{1.7}{HarnessWarm}{4.5pt}{ORIGINAL OBJECTIVE $\cdot\,\mathcal C$}
\put(49.5,37.3){\color{HarnessWarm}\rule{17.6\unitlength}{1.3\unitlength}}%
\harnessFigurePatch{50.0}{37.3}{16.8}{1.3}{HarnessWarm}{5pt}{review and follow-up}
\put(49.4,37.3){\color{HarnessBand}\rule{0.24\unitlength}{1.3\unitlength}}%
\put(49.64,37.3){\color[HTML]{9A641B}\rule{0.08\unitlength}{1.3\unitlength}}%
\put(67.0,37.6){\color{HarnessWarm}\rule{0.17\unitlength}{0.8\unitlength}}%
\put(67.16,37.6){\color{HarnessBand}\rule{0.64\unitlength}{0.8\unitlength}}
\put(67.16,37.6){\color[HTML]{9A641B}\rule{0.08\unitlength}{0.8\unitlength}}%
\put(67.8,38.03){\color[HTML]{9A641B}\vector(-1,0){0.65}}%
\put(69.5,28.22){\color{HarnessSoft}\rule{9.4\unitlength}{0.45\unitlength}}%
\harnessFigurePatch{69.5}{28.5}{9.4}{1.1}{HarnessSoft}{4.5pt}{configured model}
\harnessFigurePatch{12.1}{22.1}{20.0}{1.3}{HarnessSoft}{4.7pt}{ONE DAG NODE $\cdot$ SCOPED TASK BRIEF}
\put(92.6,40.0){\color{HarnessWarm}\rule{3.6\unitlength}{2.8\unitlength}}%
\put(94.35,41.3){\makebox(0,0){\sffamily\fontsize{5pt}{6pt}\selectfont submit}}
\end{overpic}
\caption{Execution architecture and mission-state transition. The planner converts the original objective $\mathcal C$ into scoped worker assignments. Workers append outputs and artifact references to the execution record, which review examines against the objective before requesting submission or follow-up work. Dependency edges determine dispatch order. Reviewer session initialization and benchmark-specific submission interfaces are detailed in Appendix~\ref{app:interfaces}.}
\label{fig:execution-policy}
\end{figure*}
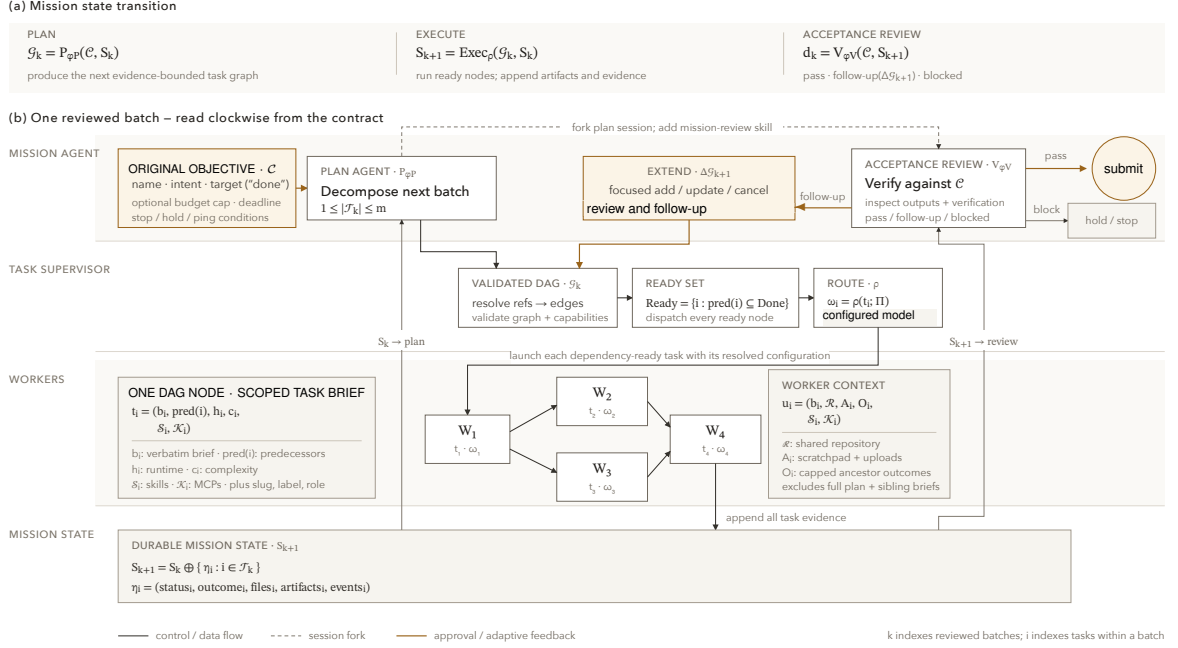

The state transition is
\begin{align}
\mathcal G_t &= P_{\varphi_P}(\mathcal C,S_t),\\
S_{t+1} &= \operatorname{Exec}_{\rho}(\mathcal G_t,S_t),\\
d_t &= V_{\varphi_V}(\mathcal C,S_{t+1}).
\label{eq:execution-transition}
\end{align}
Here $P$ and $V$ are the planning and review roles, and $\varphi_P$ and $\varphi_V$ are their procedural instructions. An \emph{agent runtime} is the execution backend, such as Claude Code or Codex, that manages a selected model's session and tool calls; the model is configured separately. The routing policy $\rho$ selects agent runtimes and applicable capabilities for the proposed work. The decision $d_t$ requests submission, further work, or stopping. Figure~\ref{fig:execution-policy} shows the objects and feedback paths.

Each graph node specifies a scoped brief, an agent runtime, applicable capabilities, and dependencies. Before execution, validation checks required fields, unique identifiers, dependency references, acyclicity, configured size bounds, and permitted agent runtimes and capabilities. Invalid submissions return actionable errors so the planner can revise the graph. Graph validation governs executable structure, while review assesses task coverage and output quality.

Dependencies define execution order: a task becomes ready when its prerequisites have completed successfully. Failed prerequisites block dependent work. Independent ready tasks may execute concurrently, and follow-up work extends the graph while preserving completed results. Benchmark-specific scheduling is described in Appendix~\ref{app:interfaces}.

\subsection{Task context and artifact handoff}

Workers receive their task brief, role instructions, and relevant task materials. Results from prerequisite tasks are passed as summaries and references to files or artifacts in the shared workspace. Workers can inspect those artifacts for additional detail, allowing each assignment to use prior results without copying complete upstream execution histories into its prompt.

The execution record links an assignment to its status, worker output, changed files, and produced artifacts. These links give review a concrete basis for inspecting the work behind a completion claim and for specifying any additional work needed. Follow-up workers can use the same recorded outputs and artifact references to continue from the existing state.

\subsection{Review, continuation, and submission}

The reviewer receives the original objective and available task results. It assesses the outputs and artifacts against the request, then either submits the result or proposes follow-up tasks with actionable briefs. Follow-up work uses the same graph validation and execution process as the initial assignments. A configured limit on continuation rounds controls how much additional work a run can request.

Benchmark adapters connect review to each domain's submission interface. HealthBench review forks the planner's session and can make minor inline answer edits or request a follow-up worker; grading uses the finalized response. In Terminal-Bench, review and repairs share the task container, whose final state is graded. Appendix~\ref{app:interfaces} details context initialization, closed-book access, and execution settings.

Official benchmark grading occurs after execution finishes. During review and continuation, the system uses the original request, available outputs, and ordinary tool feedback; official reference answers and grader verdicts remain outside this decision loop.

\subsection{Specialization during construction}

Planner, worker, and reviewer skills are reusable textual instructions. A setup agent conducts trial executions on a subset of benchmark inputs and examines the resulting outputs, role trajectories, and tool feedback to revise those instructions. For example, revisions can ask the planner to scope a clinical assignment without pre-solving it, or ask review to re-derive relevant facts from the original conversation. Official reference answers and grading feedback are excluded from this construction process. Before promotion, we manually review the instruction edits to check that they encode general procedures rather than instance-specific answers or grading information.

The finalized instructions are then used for full-benchmark evaluation, covering both inputs encountered during setup and the remaining benchmark inputs. Cross-family reuse experiments apply the same instruction content unchanged with different role models, without another setup run or model-specific retuning.

\section{Experimental Setup}
\label{sec:setup}

\subsection{Domains and native outcomes}

We evaluate complete harness configurations with their selected role models across four domains. Terminal-Bench~2.1 contains 89 command-line tasks whose official executable checks evaluate the resulting environment \citep{merrill2026terminalbench,terminalbench2026release}. HealthBench Professional (HBP) contains 525 clinician-facing conversations with task-specific physician-authored weighted rubrics \citep{hicks2026healthbenchprofessional}. ViBench evaluates applications built from product requirements using human-authored browser plans and reports Plan Pass@1 \citep{vibench2026}. DrugDiscoveryBench contains 82 scientific tasks; we report full-credit outcome pass rate under its weighted rubric judge \citep{akyurek2026drugdiscoverybench}.

These benchmarks assess distinct completion targets: an environment state, a multi-criterion clinical answer, an application workflow, and a scientific result.

\subsection{Evaluation and comparison configurations}

We report one final evaluated outcome per instance for each configuration. Full-benchmark evaluations use instructions refined in grade-blind trials and cover construction inputs and remaining tasks. Skill-reuse evaluations use these instructions unchanged with different role models, without another setup run.

Supplementary slice experiments cover Terminal-Bench~2.1, HBP, and SWE-Bench Pro \citep{deng2025swebenchpro}. Within each benchmark, comparisons use a fixed, seeded, stratified 10\% sample (seed \texttt{20260714}), with the task count rounded up: nine of 89 Terminal-Bench tasks, 53 of 525 HBP tasks, and 74 of 731 SWE-Bench Pro tasks. Stratification uses terminal category and difficulty, HBP use case, type, and difficulty, and SWE repository. The same selected identifiers are used across each benchmark's comparison arms, with benchmark-native scoring. These task-matched comparisons assess complete configurations---including role models, instructions, and inference budgets---rather than isolating planning, execution, or review. The HBP comparison covers 53 common tasks with four arms: direct Fable, Fable-configured harness, direct Opus, and Opus-configured harness. Direct HBP arms produce model completions; harness arms use planning, execution, and review. The SWE-Bench Pro slice compares the Fable reference with Fable- and Opus~4.8-configured harnesses. Appendix~\ref{app:scoring} provides benchmark-specific population details and supplementary slice results.

The full HBP system is labeled \emph{Fable-configured}. Fable~5's production safeguards can route flagged requests to Opus~4.8, including benign inputs \citep{anthropic2026fablefallback}. We preserve this fallback behavior in the evaluated configurations. Its observed root worker sessions use Fable only on 377 tasks, both Fable and Opus on 109, and Opus only on 39. Published provider and benchmark scores are presented separately from the local task-matched comparisons. The planner and reviewer are configured with Opus~4.8 across the four primary benchmark configurations; Table~\ref{tab:main-results} names their benchmark-specific worker models.

Tool access follows the benchmark environment: HBP uses the supplied conversation and local answer artifacts under a closed-book policy; terminal tasks use shell and filesystem operations in the task container; ViBench uses application code and browser-based checks; and DrugDiscoveryBench uses the supplied Biomni scientific environment. Appendix~\ref{app:interfaces} details these interfaces, network policies, and role-level reasoning-effort settings.

\subsection{Scoring and resource measurement}

Each benchmark uses its native metric. HBP computes signed rubric coverage, applies the official response-length adjustment, and clips the aggregate mean to the native score range. Appendix~\ref{app:scoring} gives the scoring formula and implementation details.

Performance, model-inference cost, and latency refer to the same evaluated instances within each configuration. Model-inference costs combine worker and planning/review usage, with model-based grading reported separately \citep{kapoor2025agentsmatter,kapoor2026hal}. Latency follows each benchmark's execution interface: HBP uses mission-event duration, excluding separate grading and provisioning; Terminal-Bench uses Harbor wall time, including preparation, execution, verification, and teardown. We report task-level mean latency. Appendix~\ref{app:resources} provides latency distributions, cost calculation details, and measurement scope.

\section{Results}
\label{sec:results}

\subsection{Cross-domain outcomes}

Table~\ref{tab:main-results} summarizes the complete configurations using benchmark-native metrics. Terminal-Bench records 83 passes and six failures (93.26\%). The 525-task Fable-configured system achieves a length-adjusted HBP score of 69.63.

\begin{table*}[t]
\centering\small
\setlength{\tabcolsep}{4pt}
\begin{tabularx}{\textwidth}{@{}l X l r r r@{}}
\toprule
Benchmark & Native metric & Harness worker & Harness & Anthropic & OpenAI\\
\midrule
Terminal-Bench 2.1 & Task pass rate & GPT-5.6 Sol & 93.26\% & 88.0\% & 91.9\%\\
HBP & Length-adjusted rubric score & Fable~5-configured & 69.63 & 66.0 & 60.5\\
ViBench & Plan Pass@1 & Opus~4.8 & 93.24\% & 87.8\% & 86.5\%\\
DrugDiscoveryBench & Full-credit pass rate & GPT-5.5 & 60.98\% & 50.04\% & 51.6\%\\
\bottomrule
\end{tabularx}
\caption{Harness evaluation results (July 2026) and published contextual references. The planner and reviewer are configured with Opus~4.8 in all four harness configurations. Anthropic references are Mythos~5 (terminal/HBP), Opus~4.8 (ViBench), and Sonnet~5 (DrugDiscoveryBench); OpenAI references are Sol Ultra, Sol, GPT-5.5, and GPT-5.5 respectively. References use their published systems and are not paired or resource-matched comparisons. The HBP harness configuration includes mixed observed worker models.}
\label{tab:main-results}
\end{table*}

Terminal and HBP references come from the model release and system cards \citep{anthropic2026fablemythos,openai2026gpt56release,openai2026gpt56}; Sol Ultra is a multi-agent system, not a single-worker Sol baseline. ViBench supplies its own references \citep{vibench2026}. DrugDiscoveryBench references use mini-SWE-agent \citep{akyurek2026drugdiscoverybench,scale2026drugdiscoveryleaderboard}. Published values provide context for the evaluations rather than a claim about the current benchmark ranking.

\subsection{Slice comparisons and paired outcomes}

Table~\ref{tab:slice-results} compares baseline execution with Fable- and Opus-configured harnesses on fixed benchmark subsets, including the previous-generation Opus~4.8 configuration. These fixed-task comparisons assess whether the harness adds value over reference execution and whether a previous-generation worker can match or exceed the newer-generation reference. The terminal slice records complete task coverage across the reference and both harness configurations: each passes all nine tasks. In the earlier reported SWE-Bench Pro slice, the Fable-configured harness achieves 82.4\% versus the Fable reference's 81.1\%, while the Opus-configured harness matches that reference at 81.1\%.

\begin{table*}[t]
\centering\small
\setlength{\tabcolsep}{4pt}
\begin{tabularx}{\textwidth}{@{}l X r r r@{}}
\toprule
Benchmark (slice) & Native metric & \shortstack{Baseline\\Fable~5} & \shortstack{Harness\\Fable-configured} & \shortstack{Harness\\Opus-configured}\\
\midrule
Terminal-Bench 2.1 (9) & Task pass rate & 100\% & 100\% & 100\%\\
HBP (53) & Length-adjusted rubric score & 52.94 & 72.83 & 59.68\\
SWE-Bench Pro (74)$^\dagger$ & Task pass rate & 81.1\% & 82.4\% & 81.1\%\\
\bottomrule
\end{tabularx}
\caption{Fixed-task slice outcomes on seeded, stratified 10\% samples (seed \texttt{20260714}; Section~\ref{sec:setup}). Parentheses give task counts; $\dagger$ denotes the previously reported SWE-Bench Pro slice. The baseline is execution without our harness, with direct model completions for HBP. Fable's safeguards can route flagged requests to Opus~4.8; the HBP Fable-configured row includes both models (Section~\ref{sec:setup}). Appendix~\ref{app:scoring} describes the populations.}
\label{tab:slice-results}
\end{table*}

Table~\ref{tab:paired-hbp} reports aggregate scores and paired task-level changes on the same 53 tasks. The Fable-configured mean increases by 19.89 points, while the median paired change is $-1.46$. The five tasks with the largest gains account for 12.83 points of the mean difference. The Opus-configured mean decreases by 1.06 points, with a median paired change of $-0.05$. Reporting both mean and median makes the concentration of gains visible alongside the typical task-level change.

\begin{table}[t]
\centering\small
\setlength{\tabcolsep}{3pt}
\begin{tabular}{@{}lrrrr@{}}
\toprule
Configuration & Direct & Harness & \shortstack{Mean\\$\Delta$} & \shortstack{Median\\$\Delta$}\\
\midrule
Fable-configured & 52.94 & 72.83 & $+19.89$ & $-1.46$\\
Opus-configured & 60.74 & 59.68 & $-1.06$ & $-0.05$\\
\bottomrule
\end{tabular}
\caption{HBP comparisons on 53 common tasks. Direct and Harness give aggregate native scores; mean and median $\Delta$ summarize paired task-level changes (Harness minus Direct), in score points. The Fable-configured arm includes mixed observed worker models.}
\label{tab:paired-hbp}
\end{table}

To distinguish rubric satisfaction from the effect of response length, we compare each harness answer with the direct answer on the same task, before and after the native length adjustment. Before adjustment, the Fable-configured harness earns more rubric points on 17 tasks, fewer on four, and the same on 32; the Opus-configured harness earns more on nine, fewer on nine, and the same on 35. After adjustment, the Fable-configured harness scores higher on 20 tasks and lower on 33; the Opus-configured harness scores higher on 24 and lower on 29. The length adjustment can therefore change the numerical ordering of answers even when their rubric scores are equal.

\subsection{Direct skill reuse}

The domain instruction content was reused unchanged with other role-model families, without another setup run. Kimi~K3 achieves 61.55 on HBP. On ViBench, GLM-5.2 achieves 71.62\%; the publicly reported GLM-family score available at the time of evaluation was 66.2\% for GLM-5.1 \citep{vibench2026}. The same instructions ran with Kimi K3 and GLM-5.2 without retuning.

\subsection{Benchmark-level resources}

\begin{table*}[t]
\centering\small
\setlength{\tabcolsep}{4pt}
\begin{tabularx}{\textwidth}{@{}Xrrrrr@{}}
\toprule
Evaluation & \shortstack{Harness\\total} & \shortstack{Harness\\per task} & \shortstack{Anthropic cost ref.\\Fable$^{*}$/Opus~4.8} & \shortstack{OpenAI cost ref.\\per task} & \shortstack{Harness\\mean latency}\\
\midrule
Terminal-Bench 2.1, Sol (89) & \$192.84 & \$2.17 & \$0.82$^{*}$ & \$5.14 & 16.94 min\\
HBP, Fable-configured (525) & \$845.10 & \$1.61 & \$0.27$^{*}$ & \$0.18 & 6.01 min\\
ViBench, Opus 4.8 (24) & \$251.24 & \$10.47 & \$2.04 & \$2.42 & 34.45 min\\
DrugDiscoveryBench, GPT-5.5 (82) & \$536.28 & \$6.54 & --- & \$2.80 & 21.79 min\\
\bottomrule
\end{tabularx}
\caption{Model-inference costs in USD; reference costs are per task. Harness costs and latency cover the same evaluated instances as the corresponding performance results; costs include planning, workers, and review. Unmarked references match the configurations in Table~\ref{tab:main-results} \citep{openai2026gpt56release,openai2026astracost,vibench2026,akyurek2026drugdiscoverybench}. ViBench costs and latency are per application build, not per browser test plan. OpenAI's terminal and HBP costs are estimated API costs. $^{*}$Anthropic cost references use Fable~5 rather than Mythos~5: terminal, 83.8\%, with mean inference cost over 445 attempts \citep{terminalbench2026fablecost}; HBP, 60.9 in OpenAI's independent evaluation. A dash denotes an unreported value.}
\label{tab:resources}
\end{table*}

Table~\ref{tab:resources} shows that the harness's cost relative to published systems depends on the benchmark and reference configuration. On Terminal-Bench, the harness spends \$2.17 per task, below Sol Ultra's \$5.14 but above Fable~5's \$0.82. On HBP, the harness spends \$1.61 per task, compared with \$0.18 for Sol and \$0.27 for Fable~5 \citep{openai2026gpt56release,terminalbench2026fablecost,openai2026astracost}. ViBench spends \$10.47 per application, compared with \$2.04 for Opus~4.8 and \$2.42 for GPT-5.5. DrugDiscoveryBench records \$6.54 per harness question against \$2.80 for the published GPT-5.5 reference. The starred Anthropic cost references use Fable~5, rather than the Mythos~5 systems whose scores appear in Table~\ref{tab:main-results}. These are contextual comparisons across published configurations, rather than resource-matched experiments. OpenAI also reports a simulated 4.28-minute latency for Sol Ultra; our terminal latency includes Harbor preparation, execution, verification, and teardown \citep{openai2026gpt56release}.

Coordination has measurable overhead in the local paired HBP comparison: mean recorded cost rises from \$0.29 to \$2.13 with Fable and from \$0.16 to \$1.16 with Opus. Corresponding mean spans rise from 0.59 to 6.80 minutes and 0.61 to 4.60 minutes. Fable-configured HBP grading costs \$98.48, separate from solving costs. Reported evaluation costs exclude setup and infrastructure. Appendix~\ref{app:resources} reports supplementary configurations, distributions, and measurement scope.

\section{Analysis}
\label{sec:analysis}

We examine paired outcomes, actual instruction revisions, and review behavior.

\subsection{Recoveries, regressions, and shared outcomes}

The paired population motivates inspecting both improvement and failure. Table~\ref{tab:cases} gives four retrospective HBP examples from the 43 paired tasks with observed Fable-only root sessions. We select the largest and smallest adjusted differences and the highest and lowest shared raw coverage; Appendix~\ref{app:cases} specifies selection. Raw rubric scores are shown to separate criterion coverage from answer length.

\begin{table}[t]
\centering\small
\begin{tabular}{lrr}
\toprule
Task type & Direct & Harness\\
\midrule
Care-instruction translation & $-100$ & 100\\
Latest clinical guideline & 100 & 0\\
Interacting conditions & 100 & 100\\
Dosing despite escalation warnings & 0 & 0\\
\bottomrule
\end{tabular}
\caption{Balanced retrospective illustrations: favorable change, regression, shared high coverage, and shared low coverage. Values are raw signed rubric scores, not binary passes or the final length-adjusted metric.}
\label{tab:cases}
\end{table}

In the translation case, the harness preserves the requested translation and flags a possible issue in the source directions. The finalized answer differs from the worker's terminal output, showing why the submission boundary matters. However, execution-summary metadata also enters the answer: a favorable grade can coexist with a defective output boundary. The regression case names an older rather than the requested latest guideline edition and produces a longer summary. For interacting conditions, added organization yields no recorded rubric gain. The shared-high and dosing cases both record consultation of the same worker instructions, but differ in their rubric outcomes.

\subsection{What specialization changes}

The setup logs show human-readable coordination changes. An earlier planner brief prescribed clinical elements to include; its revision asks the planner to frame the worker's task without pre-solving the substantive answer. Reviewer instructions move from general coverage checking toward re-deriving relevant facts from the original conversation instead of inheriting the plan's interpretation. These revisions target assumption propagation across roles.

The construction logs preserve the actual instruction edits. Selected evaluation trajectories show workers consulting domain guidance alongside the original request and answer checklist before producing an answer artifact. This makes specialization concrete: it changes readable procedures for framing, solving, and checking tasks. Appendix~\ref{app:skills} documents the instruction revisions and their recorded use.

\subsection{Review-guided verification and repair}

Terminal-Bench trajectories illustrate how review identifies errors and directs focused repairs. In \texttt{gcode-to-text}, the worker reconstructs embossed text from a printer toolpath but writes an uppercase \texttt{Z} where the rendered glyph is lowercase. The reviewer independently rasterizes the original G-code and inspects the disputed character. It requests a focused correction that preserves the remaining text and output format. The follow-up worker changes the casing and checks the final file's contents and bytes; the official verifier returns a pass. In \texttt{query-optimize}, review finds that two columns are text rather than integers despite matching printable query results. A follow-up worker adds the required casts, preserving all 500 result rows; the final artifact also passes. Appendix~\ref{app:review-repair} describes the reviewer checks and follow-up repairs for both tasks.

Other domains illustrate task-specific verification. In ViBench Notes, a verification task exercises unlock, edit, cold reload, and re-authentication; the harness artifact hides note data after reload, unlike the reference artifact. DrugDiscoveryBench checks distinguish an intact epoxide from a reacted covalent adduct and the requested neutral formula from a bound protonation state. Appendix~\ref{app:domain-examples} preserves these illustrations.

This task-dependent execution pattern is also visible at the mission level: ViBench uses one worker node in each of its 24 application missions, while DrugDiscoveryBench's 13 two-task missions add a second node only after the first terminates. The scientific follow-ups therefore represent sequential correction or recovery rather than parallel decomposition (Appendix~\ref{app:interfaces}).

The Windows virtual-machine episode provides a contrasting regression. Keyboard and reboot tests disturb a working desktop, prompting a restoration task. The repair worker restores reference framebuffer pixels but reports that the guest interface remains wedged, although final official checks pass. This contrast motivates verifying the requested functionality, not only a visual match.

Useful review converts a discrepancy into a bounded repair; narrow checks can miss functional problems, while disruptive checks can damage a working artifact, consistent with prior analyses of agent failures and self-correction \citep{cemri2025mast,huang2024selfcorrect}. Benchmark-level resource measurements account for the additional verification and repair work.

\section{Related Work}
\label{sec:related-work}

\paragraph{Execution interfaces and coordination.}
Modular model/tool architectures and reasoning--action loops established agent interfaces \citep{karpas2022mrkl,yao2023react}; ReWOO separates planning from observations \citep{xu2023rewoo}. SWE-agent, OpenHands, and Agentless show the importance of execution interfaces and structured repair \citep{yang2024sweagent,wang2025openhands,xia2025agentless}.

MetaGPT, ChatDev, and AutoGen establish role-based workflows and artifacts \citep{hong2024metagpt,qian2024chatdev,wu2024autogen}. Magentic-One uses task/progress ledgers for agent selection and recovery; WorkTeam supervises workflow construction and correction; LLMCompiler dispatches dependency-ready calls and replans \citep{fourney2024magenticone,liu2025workteam,kim2024llmcompiler}. Our protocol connects an executable work graph and artifact record to review against the original objective, with scoped continuation and benchmark-specific submission. HiAgent and StateM study context management; scaling analyses and Murakkab examine coordination costs \citep{hu2025hiagent,qin2026statem,kim2026scalingagents,chaudhry2026murakkab}.

\paragraph{Procedural specialization and harness adaptation.}
Voyager, ExpeL, Agent Workflow Memory, and SkillWeaver accumulate or refine reusable procedures \citep{wang2024voyager,zhao2024expel,wang2025awm,zheng2025skillweaver}. DSPy, MIPRO, TextGrad, and GEPA optimize programs or prompts; ADAS and AFlow search agent designs and workflows \citep{khattab2024dspy,opsahlong2024mipro,yuksekgonul2025textgrad,agrawal2026gepa,hu2025adas,zhang2025aflow}. ACE, Meta-Harness, MCE, and Agentic Harness Engineering adapt contexts or harnesses; Life Harness, SkillOpt, and Meta-Skills study runtime interfaces, skill evolution, and harness-design strategies \citep{zhang2026ace,lee2026metaharness,ye2026mce,lin2026ahe,xu2026lifeharness,yang2026skillopt,qian2026metaskills}. Our construction process supplies readable role instructions from task inputs and trajectories without official grading feedback.

\paragraph{Review and empirical evaluation.}
Self-Refine and Reflexion use model feedback; self-correction can fail without useful feedback \citep{madaan2023selfrefine,shinn2023reflexion,huang2024selfcorrect}. MAST analyzes multi-agent failures \citep{cemri2025mast}. Our comparisons and execution cases examine both corrections and regressions.

SWE-bench, Terminal-Bench, and tool--agent--user benchmarks assess operational outcomes; HBP and MDIA address clinical response evaluation and multi-agent reasoning \citep{jimenez2024swebench,merrill2026terminalbench,yao2025taubench,hicks2026healthbenchprofessional,cruz2026mdia}. AgentBoard, AI Agents That Matter, HAL, reliability studies, and log analysis motivate finer-grained evidence and resource reporting \citep{ma2024agentboard,kapoor2025agentsmatter,kapoor2026hal,rabanser2026reliability,kirgis2026loganalysis}. Our cross-domain evaluation examines these execution interfaces through task outcomes, reviewer interventions, and resource measurements.

\section{Conclusion}

An agent’s final submission depends on more than whether its workers finish their assignments. We present an execution harness that preserves the original objective throughout planning, execution, and review. Scoped assignments, artifact records, and bounded follow-up work make explicit what was attempted, produced, and assessed for submission. Reusable domain-specific instructions specialize this protocol using execution trajectories without official grading feedback.

Our evaluation across terminal tasks, clinical responses, application building, and scientific questions reveals capabilities and limitations. Paired clinical comparisons show improvements alongside regressions and persistent failures, while coordination adds cost and latency. Execution records clarify these outcomes: in a G-code decoding task, review caught a letter-casing error and directed its correction; in a virtual-machine task, additional verification disrupted a working environment. Documented instruction revisions illustrate how execution procedures can be adapted.

These findings do not imply that additional agents or review necessarily improve outcomes. Rather, they highlight submission and continuation as explicit decisions grounded in the original objective and available evidence. Evaluating these decisions alongside task outcomes and resource costs shows what the execution architecture accomplishes, where it falls short, and at what cost.

\section*{Limitations}
\label{sec:limitations}

\paragraph{System-level evaluation.}
Our experiments evaluate complete harness configurations, combining planning, execution, review, and domain-specific instructions. Isolating the contribution of each component or additional inference would require controlled ablations. The HBP and terminal evaluations primarily exercise planning, scoped execution, and review with one worker; workloads with multiple independent subtasks would provide a broader assessment of parallel coordination. Published provider results serve as contextual references for systems evaluated under their respective configurations.

\paragraph{Exposure and repeatability.}
Grade-blind construction and full evaluation can share task inputs. The evaluation therefore characterizes specialized configurations on the full benchmark population. Single evaluations and retrospective case selection do not establish repeated-run reliability or phenomenon prevalence \citep{rabanser2026reliability}.

\paragraph{Scoring and resource scope.}
Benchmark scores characterize performance under the evaluated task and grading protocols. HBP measures rubric-based response quality. Cost estimates cover worker and planning/review model usage, with grading reported separately; setup and infrastructure are excluded from these estimates. Latency follows benchmark-specific execution boundaries. Review can overlook requirements or introduce regressions, making task-aligned acceptance checks an important part of the execution policy.

\paragraph{Deployment and artifact sharing.}
Deployment beyond benchmarks calls for task-specific validation and domain-expert oversight in high-stakes clinical or scientific settings, together with appropriate safeguards for shared-workspace edits. Sharing research artifacts---including adapters, domain instructions, prompts, and execution records---should follow applicable benchmark licenses and protect sensitive information through appropriate anonymization and redaction.

\bibliography{references}
\clearpage
\appendix
\section{Native Scoring and Evaluation Sets}
\label{app:scoring}

For HBP, let $p_{ij}$ denote signed rubric points, $m_{ij}$ the saved Boolean judgment, and $L_i$ response length in characters. The native score is defined by
\begin{align}
 r_i &= \frac{\sum_j p_{ij}m_{ij}}{\sum_{j:p_{ij}>0}p_{ij}},\\
 a_i &= r_i-0.0147\frac{L_i-2000}{500},\\
 S &=100\,\operatorname{clip}\left(\frac{1}{N}\sum_i a_i,0,1\right).
\end{align}
Signed negative contributions and individual values above one remain in the mean. The coefficient and center follow the official scoring definition \citep{hicks2026healthbenchprofessional,openai2026gpt56}. Aggregate clipping follows OpenAI's \href{https://github.com/openai/simple-evals/blob/496e4570446e9f0499a00557a231a440c446df76/healthbench_eval.py}{\texttt{healthbench\_eval.py}} at commit \texttt{496e4570446e9f0499a0\allowbreak0557a231a440c446df76}.

HBP grading appends the finalized response to the supplied conversation and applies the public criterion-grading prompt in \texttt{healthbench\_eval.py}. Each rubric item receives a Boolean judgment, with up to three attempts per criterion. The reported analyses reuse these saved judgments without regenerating answers or rerunning the judge.

Task values are computed from recorded criterion judgments and response lengths, then averaged before aggregate clipping. Response length uses the recorded grader length or the saved answer's character count. Fable has 64 adjusted values below zero and 32 above one; Kimi has 80 and 39. HBP has no official binary-pass threshold in this analysis.

The Kimi-configured evaluation reports a length-adjusted HBP score of 61.55 on 525 tasks. Terminal-Bench uses its native executable result: 83/89 passes, rounded to 93.26\%.

DrugDiscoveryBench's Harbor verifier applies its \href{https://github.com/scaleapi/DrugDiscoveryBench/blob/main/benchmark/_template/tests/judge.py}{\texttt{judge.py}} to \texttt{/workspace/answer.md}, using weighted outcome rubrics. A task passes at 100\% outcome credit; process-rubric scores are separate \citep{akyurek2026drugdiscoverybench}.

The four paired HBP arms cover the same 53 identifiers. The Fable-configured harness subset has 43 Fable-only, seven mixed, and three Opus-only observed root worker model sets. A disclosed 43-task diagnostic gives 56.06 direct versus 75.22 with the harness, with a $+19.17$ mean change and $-1.46$ median. This supplementary diagnostic accompanies the complete 53-task comparison. Excluding one task with known earlier input exposure from all four arms gives mean changes of $+20.34$ and $-1.22$ on 52 tasks; the same input-exposure exclusion is applied to each arm.

The fixed slices use a deterministic, seeded, stratified draw of $\lceil 0.1N\rceil$ instances (seed \texttt{20260714}): 9/89 for Terminal-Bench, 53/525 for HBP, and 74/731 for SWE-Bench Pro. The strata are category $\times$ difficulty for terminal tasks, use case $\times$ type $\times$ difficulty for HBP, and repository for SWE. Sample membership is identical across comparison arms.

The earlier SWE-Bench Pro subset is a separate reported experiment: Fable reference 81.1\%, the harness with Fable 82.4\%, and the harness with Opus~4.8 81.1\% \citep{deng2025swebenchpro}. It is not a full-benchmark result or part of the HBP evaluation population. The nine-task Terminal-Bench comparison is 9/9 in each of its three configurations.

\section{Execution Interfaces and Scope}
\label{app:interfaces}

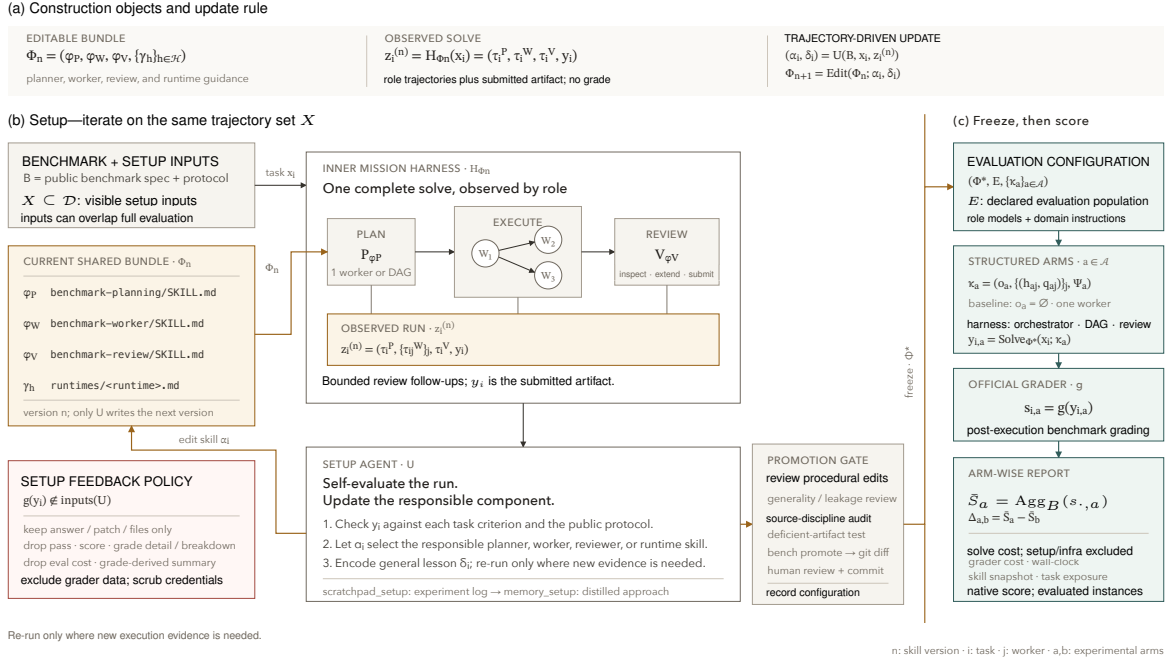
\begin{figure*}[t]
\centering
\begin{overpic}[width=\textwidth]{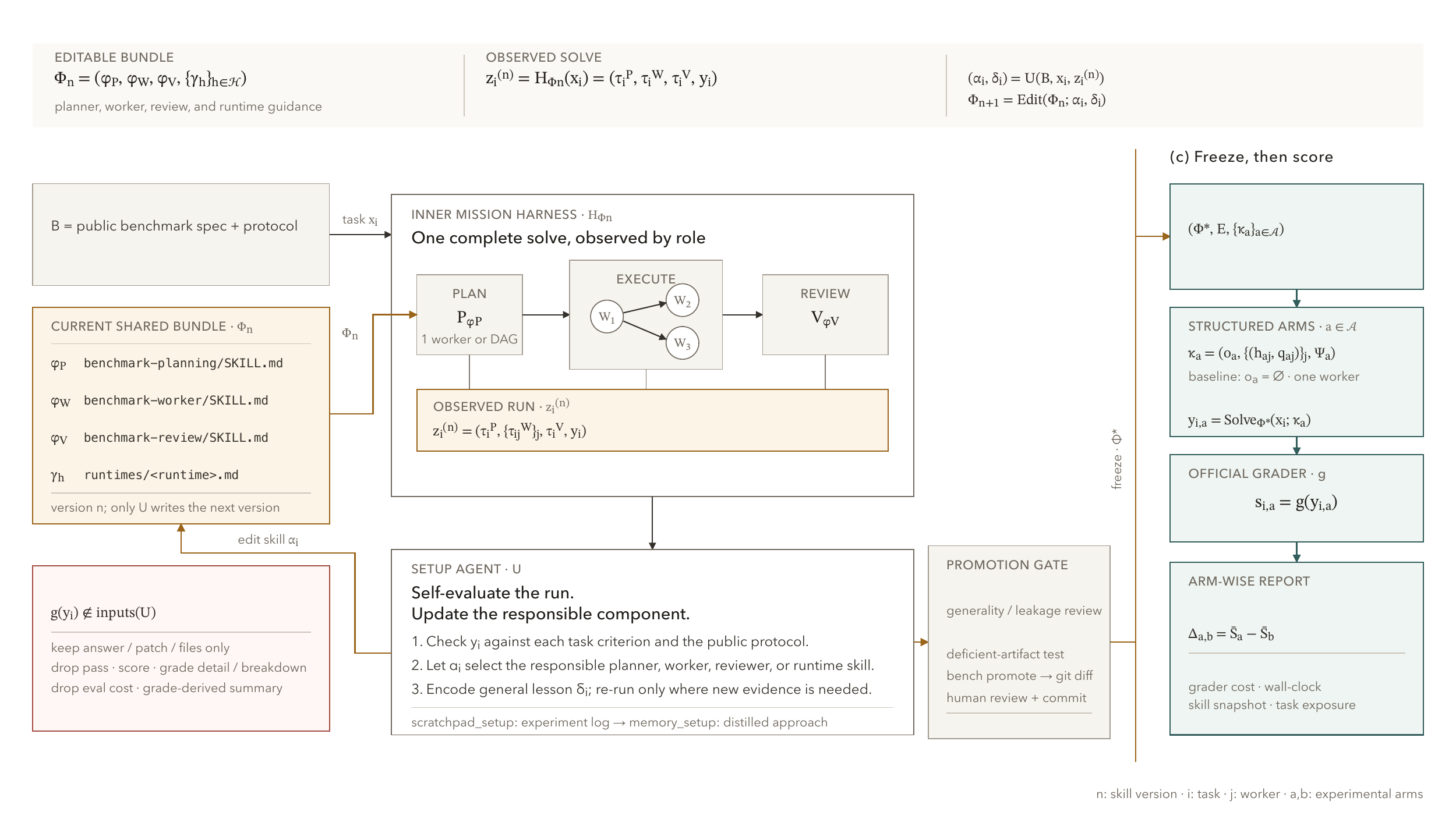}
\harnessFigurePatch{2.2}{54.05}{44.0}{1.6}{white}{5.5pt}{(a) Construction objects and update rule}
\harnessFigurePatch{33.3}{48.3}{31.0}{1.5}{HarnessBand}{4pt}{role trajectories plus submitted artifact; no grade}
\harnessFigurePatch{66.4}{51.7}{27.0}{1.5}{HarnessBand}{4pt}{TRAJECTORY-DRIVEN UPDATE}
\harnessFigurePatch{2.2}{44.8}{65.0}{1.8}{white}{5.5pt}{(b) Setup---iterate on the same trajectory set $X$}
\harnessFigurePatch{28.2}{23.35}{33.0}{1.4}{white}{4.5pt}{Bounded review follow-ups; $y_i$ is the submitted artifact.}
\put(51.7,32.25){\color{white}\rule{9.9\unitlength}{1.3\unitlength}}
\put(52.34,32.25){\color{HarnessSoft}\rule{8.66\unitlength}{1.3\unitlength}}
\put(52.34,32.25){\color[HTML]{A39B8F}\rule{0.08\unitlength}{1.3\unitlength}}
\put(60.96,32.25){\color[HTML]{A39B8F}\rule{0.07\unitlength}{1.3\unitlength}}
\put(52.45,32.25){\raisebox{0.3\unitlength}{\makebox[8.45\unitlength][c]{\color[HTML]{6E675D}\sffamily\fontsize{3.2pt}{5pt}\selectfont inspect $\cdot$ extend $\cdot$ submit}}}
\harnessFigurePatch{3.4}{41.4}{18.2}{1.6}{HarnessSoft}{5pt}{BENCHMARK + SETUP INPUTS}
\harnessFigurePatch{3.3}{38.2}{18.2}{1.8}{HarnessSoft}{5pt}{$X\subseteq\mathcal D$: visible setup inputs}
\harnessFigurePatch{3.3}{36.8}{18.9}{1.6}{HarnessSoft}{4.5pt}{inputs can overlap full evaluation}
\harnessFigurePatch{3.3}{15.0}{18.5}{1.8}{HarnessDeny}{5pt}{SETUP FEEDBACK POLICY}
\harnessFigurePatch{3.3}{6.9}{18.5}{1.3}{HarnessDeny}{4.5pt}{exclude grader data; scrub credentials}
\harnessFigurePatch{64.9}{15.3}{10.5}{1.5}{HarnessSoft}{4.5pt}{review procedural edits}
\put(64.9,5.65){\color{HarnessSoft}\rule{11.2\unitlength}{0.4\unitlength}}%
\harnessFigurePatch{64.9}{5.9}{11.2}{1.0}{HarnessSoft}{4pt}{record configuration}
\harnessFigurePatch{81.5}{41.4}{16.1}{1.7}{HarnessScore}{4.7pt}{EVALUATION CONFIGURATION}
\harnessFigurePatch{81.5}{38.2}{16.1}{1.6}{HarnessScore}{4.5pt}{$E$: declared evaluation population}
\harnessFigurePatch{81.5}{36.8}{16.1}{1.5}{HarnessScore}{4pt}{role models + domain instructions}
\harnessFigurePatch{81.5}{28.0}{16.1}{1.6}{HarnessScore}{4.2pt}{harness: orchestrator $\cdot$ DAG $\cdot$ review}
\put(81.1,19.3){\color{HarnessScore}\rule{0.5\unitlength}{1.5\unitlength}}%
\harnessFigurePatch{81.5}{19.3}{16.1}{1.5}{HarnessScore}{4.5pt}{post-execution benchmark grading}
\harnessFigurePatch{81.5}{13.4}{16.1}{1.7}{HarnessScore}{5pt}{$\bar S_a=\operatorname{Agg}_B(s_{\cdot,a})$}
\harnessFigurePatch{81.5}{9.3}{16.1}{1.7}{HarnessScore}{4.5pt}{solve cost; setup/infra excluded}
\harnessFigurePatch{81.5}{6.0}{16.1}{1.1}{HarnessScore}{4.5pt}{native score; evaluated instances}
\harnessFigurePatch{64.9}{12.05}{10.5}{1.6}{HarnessSoft}{4pt}{source-discipline audit}
\put(48.5,2.30){\color{white}\rule{50\unitlength}{1.6\unitlength}}
\harnessFigurePatch{2.2}{2.35}{41.0}{1.45}{white}{4pt}{{\color[HTML]{6E675D}Re-run only where new execution evidence is needed.}}
\end{overpic}
\caption{Grade-blind specialization as construction of domain instructions, followed by evaluation of configured systems. The schematic separates construction feedback from official grading. Full-benchmark evaluation includes construction inputs and the remaining benchmark tasks. Native aggregation is benchmark-specific, and reported resource totals exclude setup and infrastructure.}
\label{fig:setup-specialization}
\end{figure*}

\begin{table}[ht]
\centering\small
\setlength{\tabcolsep}{3pt}
\begin{tabular}{@{}lccc@{}}
\toprule
Benchmark & Planner & Worker & Reviewer\\
\midrule
Terminal-Bench~2.1 & \texttt{xhigh} & \texttt{ultra} & \texttt{xhigh}\\
HealthBench Professional & \texttt{xhigh} & \texttt{xhigh} & \texttt{xhigh}\\
ViBench & \texttt{high} & \texttt{max} & \texttt{high}\\
DrugDiscoveryBench & \texttt{max} & \texttt{xhigh} & \texttt{max}\\
\bottomrule
\end{tabular}
\caption{Role-level reasoning-effort settings for the four primary benchmark configurations.}
\label{tab:role-effort}
\end{table}

The operational objects have distinct meanings: the original objective is the completion reference; a node brief is proposed local work; dependency edges constrain dispatch; task states record execution; summaries and artifact references carry produced evidence; review supplies an acceptance or continuation judgment. Graph validation is structural. Neither a successful process exit nor a reviewer decision substitutes for official grading.

Domain specialization favors one primary worker when a task requires a coherent answer or a tightly coupled environment state. The HBP setup instructions assign one worker the complete clinical response to preserve consistency across reasoning, recommendations, and language. Terminal setup instructions likewise assign end-to-end ownership of the shared container filesystem and processes, avoiding conflicting edits and fragmented state handoffs. Additional workers are reserved for genuinely independent deliverables or focused repairs requested during review. This task-dependent policy is reflected in the evaluated graphs: HBP has 523 one-task and two two-task missions, without dependency edges; Terminal-Bench has 87 one-task and two two-task missions, with one edge. These counts refer to harness-managed nodes.

In both domains, review assesses completed work against the original objective and either concludes execution or requests focused follow-up work. The adapters provide access to the task context and artifacts through the respective execution environments.

ViBench has 24 application missions, all with one task (24 worker nodes). DrugDiscoveryBench has 82 question missions: 69 with one task and 13 with two, totaling 95 worker nodes. In all 13 two-task missions, the second task is created after the first terminates. These are sequential correction or recovery tasks, such as ``Fix protonation charge,'' ``Fix c-MYC filter,'' and ``Write answer.md,'' rather than parallel decomposition.

\paragraph{HealthBench review and submission.}
HBP initializes review by forking the planner's session; subsequent turns resume the review session. The reviewer reads the worker's answer artifact and can correct a minor factual or presentation error when preparing the submitted response, or request a follow-up worker for a substantive gap. The extension allowance defaults to two rounds, with the final turn reserved for submission. Only the finalized response is passed to the official grader.

\paragraph{Terminal-Bench review and verification.}
The terminal adapter orders worker tasks by their dependencies and executes them sequentially in one benchmark container. The reviewer inspects that same environment with the original objective and domain review instructions, returning a completion decision or focused repair briefs. Requested repairs execute in the preserved container within the configured continuation budget, including a repair requested at the final permitted round. Official checks then evaluate the final container state.

\paragraph{Tool access and network policy.}
HBP planning, worker execution, and review use the supplied clinical conversation and local files. External retrieval is disabled: the Claude-based path denies \texttt{WebSearch}, \texttt{WebFetch}, and \texttt{Bash} and omits browser tools. Terminal workers and review can read and edit files, run shell commands, and execute task-derived checks in the shared container; web access is not permitted in the Terminal-Bench~2.1 evaluation. ViBench provides an application workspace for code editing and execution, with browser-based functional checks. DrugDiscoveryBench provides the Biomni container, scientific libraries, and data resources; workers can run analyses and access permitted external scientific resources. HBP is closed-book; the other domains use their benchmark-provided environments rather than a common retrieval restriction. Planner tools submit scoped task graphs, and review inspects task outputs and either submits the result or requests follow-up work.

\section{Construction and Readable Skill Changes}
\label{app:skills}

\subsection{Construction representation}

Figure~\ref{fig:setup-specialization} outlines the construction process for specializing role instructions. Its editable instructions are represented by
\begin{equation}
\Phi_n=(\varphi_P,\varphi_W,\varphi_V,\{\gamma_h\}_{h\in\mathcal H}),
\end{equation}
where $\gamma_h$ denotes agent-runtime guidance when present. A solve yields an artifact and available role trajectories:
\begin{equation}
z_i^{(n)}=H_{\Phi_n}(x_i)=(\tau_i^P,\tau_i^W,\tau_i^V,y_i).
\end{equation}
The setup role $U$ selects a procedural revision $\delta_i$ to component $\alpha_i$:
\begin{align}
(\alpha_i,\delta_i)&=U(B,x_i,z_i^{(n)}),\\
\Phi_{n+1}&=\operatorname{Edit}(\Phi_n;\alpha_i,\delta_i).
\label{eq:setup-update}
\end{align}
These equations specify how the setup role revises procedural instructions from execution records. Input exposure and benchmark-native aggregation are documented in Appendix~\ref{app:scoring}.

The documented HBP and terminal construction runs each use 25 trial executions, including reruns to check instruction revisions. The setup agent examines task inputs, execution trajectories, tool feedback, and produced artifacts to refine the role instructions, without official answer keys or grader verdicts. Full evaluation covers both inputs encountered during construction and the remaining benchmark tasks.

The setup skill instructs the agent to inspect the final output and execution trajectory, identify the responsible role, and encode general procedural lessons in its instructions. Repeated executions check whether a revision addresses the observed issue while preserving useful behavior. The construction records include both the instruction edits and the setup agent's final summary of the chosen execution approach.

\begin{table*}[t]
\centering\small
\begin{tabularx}{\textwidth}{@{}l X X@{}}
\toprule
Role & Earlier instruction / procedure & Revised instruction / procedure\\
\midrule
HBP planner & \emph{understand the medical task, then hand it to one worker} & \emph{frame the task for one worker---do not solve it yourself}\\
HBP reviewer & General accuracy and coverage checking & \emph{Re-derive the key clinical facts yourself} from the source conversation, rather than the plan's framing\\
Terminal worker & Orient before acting & Produce a complete runnable artifact at the requested path early, then iterate\\
Terminal reviewer & General completion checking & Programmatically test machine-checkable requirements rather than rely on visual plausibility\\
\bottomrule
\end{tabularx}
\caption{Procedural revisions recorded in the setup logs. Italicized phrases are instruction excerpts; other cells paraphrase the changes. Chronology and write success do not identify an isolated performance effect.}
\label{tab:skill-edits}
\end{table*}

Two documented procedural revisions illustrate specialization. \emph{Planner:} an earlier clinical brief prescribed substantive elements to include; the later version asks the planner to scope the assignment without pre-solving the answer. \emph{Reviewer:} general coverage checks become instructions to re-derive relevant facts from the original conversation instead of accepting the plan's interpretation. These are readable coordination changes, not task-answer insertions. The archived setup conversations preserve both instruction versions. The shared-high and shared-low HBP cases in Section~\ref{sec:analysis} both record loading of the promoted worker instructions and consultation of the original task and answer checklist.

The reported cross-family experiments directly reuse the same domain instruction content without another setup run or model-specific skill retuning. This is an artifact-reuse observation rather than an isolated transfer-effect estimate.

\subsection{Setup-agent process skill}
\label{app:setup-skill}

The setup agent loads a process skill before revising domain-specific role instructions. Appendix~\ref{app:setup-full} reproduces the complete skill, including role definitions, tool access, grade-blind constraints, and the iterative revision procedure.

\section{Retrospective Case Selection}
\label{app:cases}

The four HBP cases come from the 43 paired tasks with observed Fable-only root worker model sets. Selection uses the largest and smallest adjusted difference, highest shared raw coverage, and lowest shared raw coverage. Ties use absolute adjusted difference and then lexical task identifier. This is declared outcome-informed selection, not a frequency estimate. Clinical descriptions are de-identified paraphrases and saved rubric judgments are not independent medical adjudications.

The translation case also exposes an output-boundary problem: finalization adds material beyond the worker's terminal output, including execution-summary metadata. The shared-low case demonstrates that explicit skill ingestion does not eliminate escalation-related failure. In the terminal VM case, review disturbs the environment, restoration matches reference pixels, and final checks pass despite reported guest-interface problems.

\subsection{Reviewer-guided corrections on terminal tasks}
\label{app:review-repair}

These examples illustrate reviewer-guided repairs on Terminal-Bench tasks. The G-code input and SQL reference query match the corresponding TB2.1 artifacts byte-for-byte. These case studies are separate from the 89-task aggregate reported in Table~\ref{tab:main-results}.

\begin{figure*}[t]
\centering
\includegraphics[width=.46\textwidth]{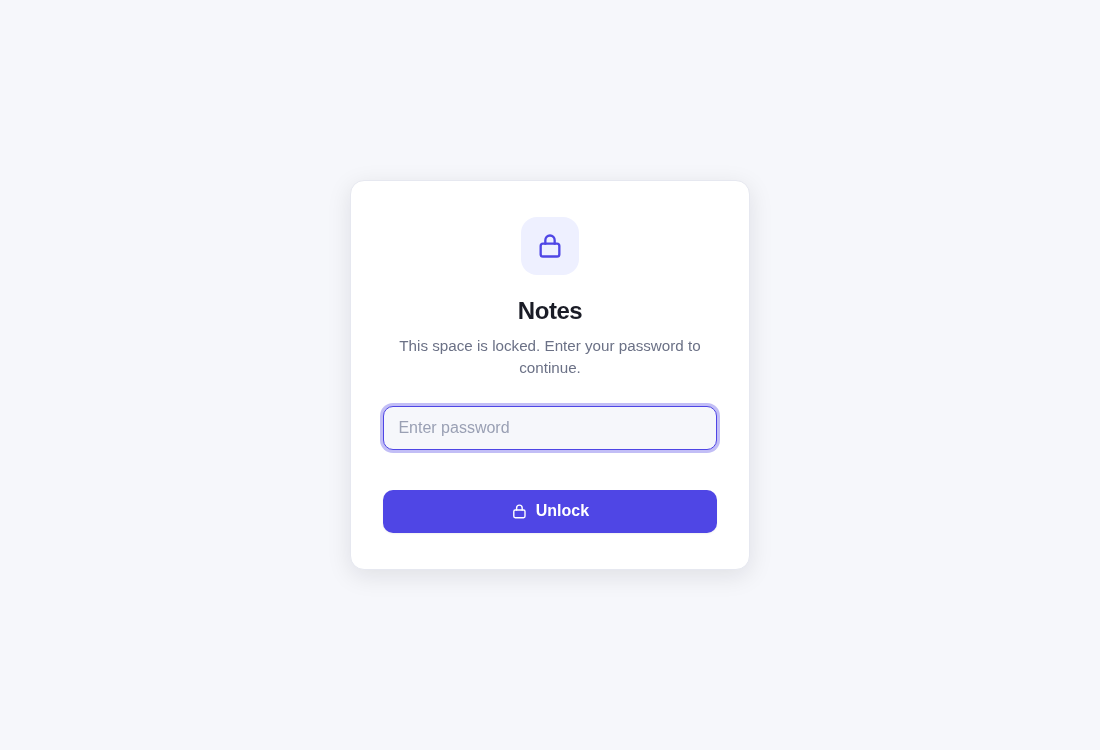}\hfill
\includegraphics[width=.46\textwidth]{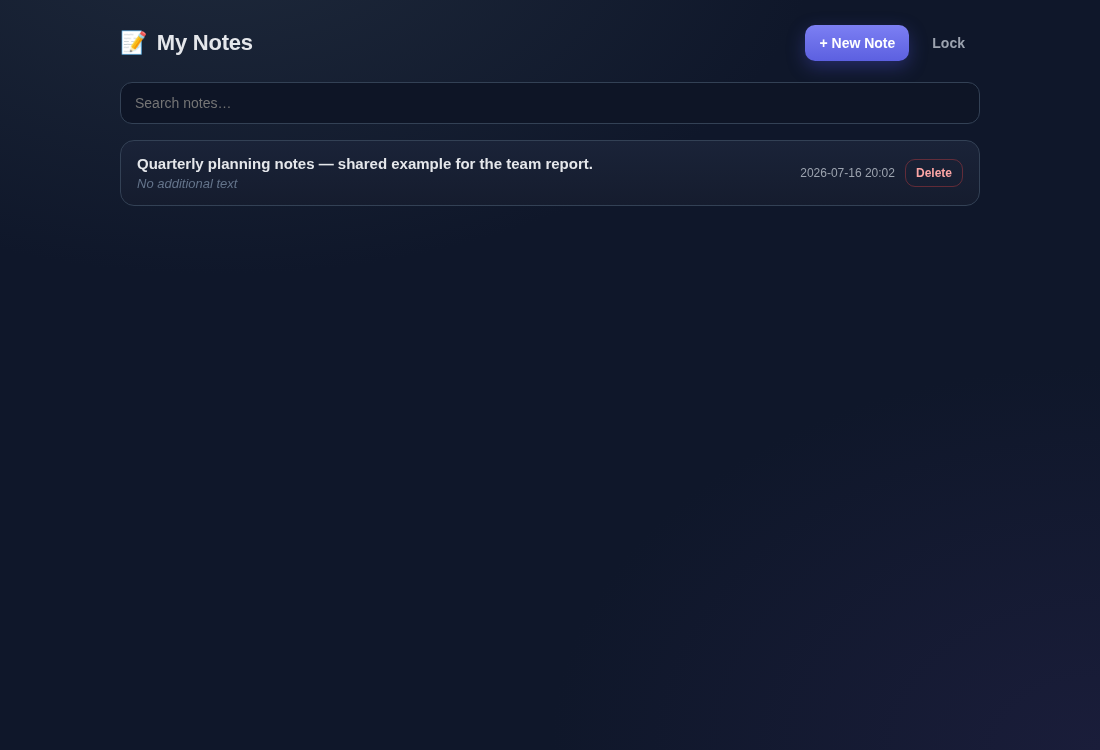}
\caption{ViBench Notes artifacts after a cold reload. The harness returns to a password gate with note data hidden (left), whereas the comparison artifact remains unlocked (right). This is a task-specific state observation, not the aggregate Plan Pass@1 measure or a component ablation.}
\label{fig:vibench-reload}
\end{figure*}

\subsubsection*{Source-grounded text correction.}
In \texttt{gcode-to-text}, the requested artifact is \texttt{/app/out.txt}, containing the text embossed by a supplied printer toolpath. The initial worker writes \texttt{flag\{gc0d3\_iZ\_ch4LLenGiNg\}}. The reviewer reads the original G-code, independently rasterizes the extrusion moves belonging to the text object, and zooms into the disputed glyph. It identifies a lowercase \texttt{z}, consistent with the worker's own readable rendering. Its \texttt{complete:false} decision creates one focused follow-up task: correct that character while preserving the other letters and output format. The repair worker updates the file to \texttt{flag\{gc0d3\_iz\_ch4LLenGiNg\}}, removes a newline introduced by the edit, and verifies its contents and 26-byte length. The final official task result is a pass.

\subsubsection*{Typed query equivalence.}
In \texttt{query-optimize}, workers produce an optimized SQLite query with 500 printable result rows matching the supplied reference. The reviewer checks both queries on a database copy and inspects output types. Two final columns, obtained through \texttt{substr}, are text rather than the reference query's integers. It requests only two \texttt{CAST(... AS INTEGER)} wrappers, preserving the query logic and database. The follow-up worker applies them, verifies integer types across all 500 rows, and confirms unchanged printable output. The final official task result is a pass.

\begin{table}[ht]
\centering\small
\setlength{\tabcolsep}{3pt}
\begin{tabular}{@{}lcc@{}}
\toprule
Artifact property & Before repair & After repair\\
\midrule
G-code text segment & \texttt{iZ} & \texttt{iz}\\
G-code output bytes & 26 & 26\\
SQL final-column types & Text & Integer\\
SQL printable result rows & 500 & 500\\
\bottomrule
\end{tabular}
\caption{Recorded artifact changes following reviewer-created tasks. Both final task outcomes pass; the table compares local artifact properties rather than intermediate official grades.}
\label{tab:review-repair}
\end{table}

The reviewer derives each correction from the original input or reference artifact and turns it into a scoped continuation. These sequences show useful review beyond trusting an apparently completed worker result.

\section{Additional Domain Examples}
\label{app:domain-examples}

The application verification task exercises unlock, edit, reload, observation, and re-authentication from an empty database. Its acceptance target is the post-reload authentication boundary. In the scientific examples, checks distinguish an intact epoxide in PDB 8B0E from a reacted adduct and a requested neutral molecular formula from the bound protonation state in PDB 2PIY. These original domain illustrations show how a procedural check can refer to the intended representation or behavior without embedding an instance answer in reusable instructions.

\section{Resource Distributions and Reproduction}
\label{app:resources}

\begin{table}[ht]
\centering\small
\begin{tabular}{lrrr}
\toprule
Evaluation & Mean & Median & p90\\
\midrule
HBP, Fable-configured & 6.01 & 5.37 & 8.14\\
HBP, Kimi-configured & 8.85 & 7.73 & 13.85\\
Terminal-Bench 2.1 & 16.94 & 13.77 & 31.79\\
\bottomrule
\end{tabular}
\caption{Recorded task latency distributions, in minutes. HBP uses mission-event spans; terminal uses Harbor wall time.}
\label{tab:resource-detail}
\end{table}

ViBench covers 24 application builds (\$251.24 total inference cost; \$10.47 per application), with 34.45-minute mean latency. DrugDiscoveryBench covers 82 questions (\$536.28 total; \$6.54 per question), with 21.79-minute mean latency. ViBench latency includes solving and browser-based grading; DrugDiscoveryBench latency spans the Harbor trial through verification.

The supplementary Kimi-configured HBP evaluation covers 525 tasks: \$661.14 in total model-inference cost, \$1.26 per task, and 8.85-minute mean latency. Separate model-grader expenditure totals \$98.36 for Kimi and \$98.48 for Fable.

Worker and planning/review spend respectively total \$493.67/\$351.44 for Fable-configured HBP, \$191.58/\$469.57 for Kimi-configured HBP, and \$97.15/\$95.70 for Terminal-Bench. Rounded components can differ from displayed totals.

Performance, cost, and latency refer to the same evaluated instances within each configuration. Totals sum unrounded per-instance SDK-reported or model-priced usage, including workers and planning/review but excluding setup and infrastructure. The 53-task Fable-configured subset reuses full-evaluation missions and is counted only once. Terminal agent execution averages 15.88 minutes within the 16.94-minute Harbor interval; these intervals are not additive. Per-task latency distributions do not predict parallel batch wall time.

Implementation checks and synthetic tests cover signed rubric points, aggregation order, response-length handling, population membership, unclipped paired differences, model attribution, and evidence-link consistency. Numeric summaries and execution examples support the analysis. Public release of harness adapters, setup prompts, initial/final instructions, or recorded trajectories would follow applicable licensing and privacy requirements, with sensitive content redacted.

\section{Complete Setup-Agent Skill}
\label{app:setup-full}
The following panels reproduce the complete setup-agent process skill. Identifying system names are anonymized and line wrapping is adapted for presentation; the substantive instructions are unchanged.

\begin{figure*}[p]
\centering
\begingroup
\setlength{\fboxsep}{8pt}
\setlength{\fboxrule}{0.4pt}
\begin{lrbox}{\setupSkillBox}
\begin{minipage}{\dimexpr\textwidth-2\fboxsep-2\fboxrule\relax}
\raggedright
{\sffamily\bfseries Setup-agent process skill\par}
\smallskip
\begin{Verbatim}[fontsize=\normalsize,breaklines=true,breakanywhere=true,breaksymbolleft={},breaksymbolright={},breakautoindent=true]
---
name: setup
description: REQUIRED first read for the benchmark-harness setup agent. The
process for authoring and iteratively improving a benchmark's harness skills
(benchmark-planning / benchmark-worker / benchmark-review) by running
instances, diagnosing what the planner/worker/reviewer agents got wrong, and
folding GENERAL lessons back into the skills — driving them toward a global
optimum that generalizes to the hidden test set. Experiment one approach at a
time, never in parallel. Follow this exactly.
---

# benchmark-harness setup agent
You author and tune the **harness skills** for one benchmark. Those three
skills — `benchmark-planning` (the planner), `benchmark-worker` (the worker),
`benchmark-review` (the reviewer) — are the instructions the real harness
**agents** follow when they solve this benchmark. Your job is to make them
good: run harness on real instances, find where the planner / worker / reviewer
**agents** got it wrong, and fold a **general** lesson back into the right
skill so it stops making that class of mistake. Keep folding lessons in and
iterating **until you have real conviction in your approach** — then move on.

**Your ultimate objective.** These three skill files are integral to harness's
architecture, but the goal is not to make harness do well on the instances
*you* run. It is to drive these three skills toward a **global optimum** —
general instructions that make the planner / worker / reviewer perform well
when harness is **scaled to the full, hidden test set you never see**.
Everything below serves that: judge by your own understanding (not a grader),
keep every lesson general (never instance- or grader-specific), and prefer the
approach that will generalize — not the one that happens to fit today's batch.

You are **not** solving the benchmark tasks yourself. You are writing the
playbook harness uses after the skills are finalized.

## What each role means — and the simplest-first rule
Harness solves an instance as a small team of **separate agents**, and your
three skills are the instructions each of those agents runs on. Before you
author anything, **understand these roles for real** — from the descriptions
below and from the actual planner/worker/reviewer **trajectories** you read via
`locate_trajectory`, so you know exactly what you're refining. (You are
sandboxed to your own working dir and cannot read the benchmark's source —
that's deliberate; see the grade-blind rule.) The three roles:

- **`benchmark-planning` (planner)** — its **own agent**. It first
  **understands the task end-to-end** and improves its own understanding of
  what's being asked, and only THEN decides how to attack it: how many workers,
  and the clear **objective + task info** it hands each one. The plan is **not
  always a multi-worker DAG** — decompose into multiple workers *only* when the
\end{Verbatim}
\end{minipage}
\end{lrbox}
\fcolorbox{black!30}{black!3}{\usebox{\setupSkillBox}}
\endgroup
\caption{Complete setup-agent skill.}
\label{fig:setup-full-1}
\end{figure*}
\clearpage

\begin{figure*}[p]
\centering
\begingroup
\setlength{\fboxsep}{8pt}
\setlength{\fboxrule}{0.4pt}
\begin{lrbox}{\setupSkillBox}
\begin{minipage}{\dimexpr\textwidth-2\fboxsep-2\fboxrule\relax}
\raggedright
{\sffamily\bfseries Setup-agent process skill (continued)\par}
\smallskip
\begin{Verbatim}[fontsize=\normalsize,breaklines=true,breakanywhere=true,breaksymbolleft={},breaksymbolright={},breakautoindent=true]
  task has genuinely independent parts worth doing in parallel and then
  synthesizing. Decomposition for its own sake usually **hurts** coherence and
  accuracy, so the planner must justify a split from the task's structure,
  never reach for one by reflex. (The simplest-first bias below is what tells
  the planner to start with one worker.)
- **`benchmark-worker` (worker)** — its own agent; **this is where the heavy
  work happens**: it actually **does the task it was handed** (writes the
  answer / the patch) and **verifies its own work before finishing**. Whoever
  produces the **final deliverable** must **persist it to the filesystem** —
  the exact artifact the reviewer reads. In a single-worker plan that's this
  worker; in a **decomposed DAG** it's the **synthesis / terminal worker** that
  owns and writes the final result, while upstream workers hand their parts
  downstream. Whoever owns the final answer, "chatting" it instead of writing
  the agreed file breaks the handoff to the reviewer. This skill should carry
  **all the benchmark-specific context** the worker needs to be effective — the
  domain, the output format and conventions, the tools/environment it works in,
  and what a good result looks like — so the worker is fully **contextual** on
  its own, without depending on the planner to re-explain the benchmark each
  time.
- **`benchmark-review` (reviewer)** — its own agent and the **last gate before
  submit**; a lot of the quality is won or lost here, so it must be **strict,
  not a rubber-stamp**. It reviews the worker's **actual written output**
  (reads the artifact the worker wrote and reviews THAT) against what
  **"correct" really requires** for this benchmark. Tools: a **task digest** +
  **`get_task(<id>)`** to see what each worker did and **which file it wrote**
  (so it can tell a draft from the synthesis worker's final answer, and read
  THAT file). **During review, be critical**: decompose the answer against the
  benchmark's evaluation criteria and actively hunt for defects — a missing
  required element, an unsupported / over-claimed assertion, something the task
  asked for that isn't there, an incomplete set. **On any real gap, drive an
  improvement rather than shipping it — commission a follow-up worker with
  `submit_task_graph`** (name the specific gap; one writer per scratchpad file)
  and **re-review**; a small mechanical slip it can fix inline. This critical,
  iterative improvement *before* the answer is final is exactly **where the
  reviewer adds quality** — and it does so by **commissioning work, not by
  rebuilding the answer itself**. Only the **final** step is hands-off: once
  review is genuinely done and the answer is good,
  **`submit_benchmark_result`** it **without rewriting it from scratch** —
  re-deriving a finished good answer degrades it (dropped content, padded
  length, baked-in mistakes). **A reviewer that keeps submitting drafts
  verbatim ("nothing to fix") is under-reviewing, not doing no harm** — the
  do-no-harm rule protects a genuinely good *final* answer, it is **never**
  license to rubber-stamp a deficient one. Rounds are **bounded** (its prompt
  says which) — out of rounds, fix any clear accuracy/safety defect inline and
  submit, since submitting beats submitting nothing. Where the benchmark allows
  (e.g. code) the checks include **writing hard tests that mimic the hidden
  ones and running them**, and **checking for regressions** — run the
\end{Verbatim}
\end{minipage}
\end{lrbox}
\fcolorbox{black!30}{black!3}{\usebox{\setupSkillBox}}
\endgroup
\caption{Complete setup-agent skill (continued).}
\label{fig:setup-full-2}
\end{figure*}
\clearpage

\begin{figure*}[p]
\centering
\begingroup
\setlength{\fboxsep}{8pt}
\setlength{\fboxrule}{0.4pt}
\begin{lrbox}{\setupSkillBox}
\begin{minipage}{\dimexpr\textwidth-2\fboxsep-2\fboxrule\relax}
\raggedright
{\sffamily\bfseries Setup-agent process skill (continued)\par}
\smallskip
\begin{Verbatim}[fontsize=\normalsize,breaklines=true,breakanywhere=true,breaksymbolleft={},breaksymbolright={},breakautoindent=true]
  **existing test suite** and exercise related functionality to confirm the
  change didn't break something already present in the codebase (a fix that
  passes the new test but regresses existing behavior still fails). (If the
  worker's output isn't where the reviewer expects it, that's a **handoff bug
  to fix in the skills**.) How heavy review should be is
  **benchmark-dependent** — decide its weight from what "correct" actually
  requires.

**Start with the simplest approach that could work — a single worker — and add
structure only when a test shows it helps.** This bias lives in the setup
process and shapes what the *planner* skill tells the planner to do (start with
one worker; only escalate when warranted). It's a starting point, not a
verdict: you then EXPERIMENT (below) to find the right level for THIS
benchmark. Absolutely-parallel work can split-then-synthesize; most cannot.

## Experiment, don't assume
You do not know up front whether single-worker, decomposition, or a
multi-perspective approach wins for this benchmark — you **find out by
testing**. Treat each strategy as an **approach** you run as an experiment:

1. **Hypothesize** — in `scratchpad_setup.md`, write the approach you're about
   to test and what you expect (e.g. "single worker will be more coherent than
   a 3-way split here").
2. **Test on a FIXED instance set** — run the *same* instances under the
   approach. Comparing approaches means re-running the same instances under
   each — that is not waste, it **is** the experiment.
3. **Measure** — from the trajectory + output (never a grade): did the behavior
   change as you predicted? is the output better by your judgment? Record the
   verdict in the scratchpad.
4. **Adopt or discard** — fold the winning approach into the skills; drop what
   didn't help. Keep the **simplest** approach that wins.

**Experiment one approach at a time — NEVER in parallel.** Test an approach
**fully** before you move to the next: encode it in the skills, run your fixed
instance set, do all the re-runs it needs, and reach a verdict — *then* change
the skills to the next approach and repeat. Running two approaches at once is
**incorrect and impossible to attribute**: all runs read the SAME skill files
on disk (one shared override dir), so whatever the skills currently say is what
every in-flight run uses. There is no way to have instance A on "single worker"
and instance B on "decompose" simultaneously — they'd both use the one skill
state on disk. Sequential is the only correct way to compare approaches.

## Your task inputs
Your task message gives you, concretely:

- **`benchmark`** — the benchmark id (e.g. `healthbench-professional`,
  `swe-bench-pro`, `terminal-bench`).
\end{Verbatim}
\end{minipage}
\end{lrbox}
\fcolorbox{black!30}{black!3}{\usebox{\setupSkillBox}}
\endgroup
\caption{Complete setup-agent skill (continued).}
\label{fig:setup-full-3}
\end{figure*}
\clearpage

\begin{figure*}[p]
\centering
\begingroup
\setlength{\fboxsep}{8pt}
\setlength{\fboxrule}{0.4pt}
\begin{lrbox}{\setupSkillBox}
\begin{minipage}{\dimexpr\textwidth-2\fboxsep-2\fboxrule\relax}
\raggedright
{\sffamily\bfseries Setup-agent process skill (continued)\par}
\smallskip
\begin{Verbatim}[fontsize=\normalsize,breaklines=true,breakanywhere=true,breaksymbolleft={},breaksymbolright={},breakautoindent=true]
- **`skillsDir`** — the directory where you AUTHOR the benchmark skills. Write
  files here:
  - `benchmark-planning/SKILL.md`
  - `benchmark-worker/SKILL.md`
  - `benchmark-review/SKILL.md`
  - `runtimes/<runtime>.md` (per-runtime guidance, only if the benchmark uses
    it)
    Files you write here are picked up by the very next run — no rebuild.
- **`scratchpadFile`** — `scratchpad_setup.md`. Your LIVE experiment log for
  THIS session (see below).
- **`memoryFile`** — `memory_setup.md`. The distilled, durable record you write
  at the END (see below).

## Tools
Benchmark control (in-process `benchmark-setup` tools):

- **`list_instances`** — the instances available to you (ids + light tags).
  This is your working set; run and diagnose only these.
- **`describe_instance`** — one instance's task definition, exactly what the
  real solver sees (the prompt/conversation; for code benchmarks the problem
  statement + repo; etc.), plus the public labels the dataset ships (e.g.
  use_case / type / difficulty, repo, and so on — whatever that benchmark
  provides). It does **not** expose the grader's hidden criteria at all —
  whatever this benchmark grades against (rubric text,
  fail-to-pass/pass-to-pass test names, etc.), and **not even their counts**.
  Use it to pick a **relevant, diverse** batch — not to memorize answers.
- **`run_instances`** — run an explicit list of instance ids with the CURRENT
  authored skills. Returns a `runId` and, per instance, the **live trajectory
  paths**. **Runs are expensive** (see Cost discipline).
- **`run_status`** — per-instance INFRA status (pending / running /
  graded=finished / error / cancelled)
  - progress. `graded` means the cell FINISHED, **not** that it passed. Check
    this to know when a run is done; don't guess.
- **`wait`** — whenever runs are in progress and you find yourself waiting for
  something (instances to finish, a trajectory to reach a point), call
  `wait(minutes)`: it pauses you and resumes you automatically after `minutes`
  (1–5, clamped; default 1) while the run keeps progressing in the background.
  **Pick the duration for the situation:** wait **longer (3–5 min)** when a
  full initial run is churning and there's nothing to do until it finishes;
  wait **shorter (1–2 min)** when you're tailing a live trajectory to confirm a
  specific fix and want to check back soon. Pass `runId` to get the status back
  on wake. This is purely to manage tokens/cost — it stops you from burning
  turns polling in a loop. Call it as many times as you need; nothing must
  follow a single `wait`.
- **`get_output`** — the OUTPUT harness produced per instance (e.g. the Q&A
  answer for HBP, the patch size for code benchmarks, etc. — whatever the
  deliverable is). This is the artifact **you** judge — it deliberately does
\end{Verbatim}
\end{minipage}
\end{lrbox}
\fcolorbox{black!30}{black!3}{\usebox{\setupSkillBox}}
\endgroup
\caption{Complete setup-agent skill (continued).}
\label{fig:setup-full-4}
\end{figure*}
\clearpage

\begin{figure*}[p]
\centering
\begingroup
\setlength{\fboxsep}{8pt}
\setlength{\fboxrule}{0.4pt}
\begin{lrbox}{\setupSkillBox}
\begin{minipage}{\dimexpr\textwidth-2\fboxsep-2\fboxrule\relax}
\raggedright
{\sffamily\bfseries Setup-agent process skill (continued)\par}
\smallskip
\begin{Verbatim}[fontsize=\normalsize,breaklines=true,breakanywhere=true,breaksymbolleft={},breaksymbolright={},breakautoindent=true]
  **not** include the grader's pass/fail or score (see rule 3).
- **`locate_trajectory`** — returns a **path** (never dumps content) to a
  role's trajectory (`planner` / `worker` / `reviewer`) for an instance, live
  while running or persisted once done. Read it yourself with Read/Grep/tail so
  you control your own context.
- **`cancel`** — stop a run (or let it finish). Use it for early-exit (see Cost
  discipline).

Native tools: **Read / Write / Edit** (author skills + memory), **Grep /
Glob**, **Bash** (shell utilities inside your working dir + trajectory dirs —
you are OS-sandboxed and **cannot** read the benchmark's grader/adapter/harness
source or the runs db; that is intentional, see the grade-blind rule),
**WebSearch / WebFetch** (research the benchmark from public docs — but never
look up the specific per-instance grading rubric / answer keys for the
instances you run, even if public). **Task / subagents** — your parallelism
engine: fan out one subagent per failing instance (or per failure cluster) to
diagnose concurrently; then you implement the fixes and integrate the general
instructions that improve the skills.

## Rules that cannot be broken
1. **Never leak dataset OR grader specifics into the skills.** The skills must
   contain only **general** instructions and principles that help on ANY
   instance of this benchmark — never an instance id, a specific expected
   answer, a specific test name, a specific repo/file, or anything that only
   helps the instances you happened to run. Equally, never encode
   **evaluation/grader/judge specifics** — no rules written to satisfy a
   particular rubric item, judge wording, or grader heuristic; that games the
   metric instead of genuinely improving the work. If a lesson only makes sense
   for one instance or one grader quirk, it is overfitting — generalize it to
   the underlying principle or drop it. The skills ship against the whole set;
   instance- or grader-specific text is cheating and will not generalize.

2. **Judge quality yourself — there is no grader verdict to use.** The runs you
   commission are **grade-blind by construction**: the harness skips grading
   them and stores **no** verdict of any kind — no pass/fail (e.g. SWE /
   terminal-bench), no score (e.g. HBP), whatever this benchmark grades on — so
   there is nothing in the db to look up, even with Bash. This is deliberate:
   tuning skills to the grader's verdict games the eval instead of genuinely
   improving the work. Decide whether harness did the task correctly from the
   **final output + trajectory + tool calls**, using your own understanding of
   what a correct solution looks like. **Judge the way the hidden grader would,
   not with a holistic "is this a strong answer?" gut feel — that vague check
   is exactly how real defects survive.** Decompose quality into the criteria
   THIS benchmark actually rewards and penalizes (inferred from the public
   rubric structure / axes, the task's stated goal, and what a domain expert
   would demand), and test each output against **each** criterion
   **adversarially** — actively hunt for what's missing, wrong, unsupported,
\end{Verbatim}
\end{minipage}
\end{lrbox}
\fcolorbox{black!30}{black!3}{\usebox{\setupSkillBox}}
\endgroup
\caption{Complete setup-agent skill (continued).}
\label{fig:setup-full-5}
\end{figure*}
\clearpage

\begin{figure*}[p]
\centering
\begingroup
\setlength{\fboxsep}{8pt}
\setlength{\fboxrule}{0.4pt}
\begin{lrbox}{\setupSkillBox}
\begin{minipage}{\dimexpr\textwidth-2\fboxsep-2\fboxrule\relax}
\raggedright
{\sffamily\bfseries Setup-agent process skill (continued)\par}
\smallskip
\begin{Verbatim}[fontsize=\normalsize,breaklines=true,breakanywhere=true,breaksymbolleft={},breaksymbolright={},breakautoindent=true]
   over-claimed, or not-what-was-asked, rather than confirming it looks fine.
   You are the **proxy for the reviewer you author**: if you rubber-stamp
   outputs, the reviewer skill you write will rubber-stamp too.

3. **Runs are expensive — spend them deliberately** (Cost discipline, below).

## Cost discipline
Every instance run costs real time and money (some pull multi-GB images and run
for many minutes).

- **Run budget: 25 instance-runs per APPROACH, 100 per session.** Each distinct
  approach you test gets its own budget of ~25 (its initial runs + the re-runs
  that verify it); across all approaches you may run at most **100** this
  session (so ~4 approaches). Plan across them; track every run + which
  approach it belonged to in your `scratchpad_setup.md` ledger so you never
  lose count or re-run one by accident. (There is no hard code cut-off — this
  budget is yours to respect; spend is audited after.)
- **Initial batch: ~10 instances, deliberately diverse and distinct** from one
  another (different repos / task types / failure shapes — not near-duplicates
  of the same case). Spread across the documented axes (category × type ×
  difficulty × language, whatever this benchmark ships) and deliberately
  include the **hard / adversarial slice** (e.g. red-teaming, difficult) where
  skills are most likely to break — not just easy typical cases. This first
  batch should give you broad signal — and it becomes the FIXED set you reuse
  to compare approaches.
- Within ONE approach, don't re-run an instance whose output already looks
  right to you. **Across approaches, re-running the same instances is the
  point** — it's how you compare strategies fairly.
- **On an INITIAL run (before you've changed anything), let instances run to
  completion** so you can read the FULL output harness produced (`get_output` +
  `locate_trajectory`) and judge what's weak. **Early-cancel is only for
  TESTING a fix**: once the live trajectory confirms the behavior changed (e.g.
  the worker now runs the repo's tests before finishing), you usually don't
  need the full solve — `cancel` and move on.
- Fan out diagnosis with subagents (cheap); reserve `run_instances` for when
  you actually need fresh evidence.

## The loop
1. **Bootstrap the v0 skills (simplest-first).** Research the benchmark from
   what the real solver sees (`describe_instance`) + its public description on
   the web: its task, what "done" means, closed-book vs tools. **Always
   web-search whether this benchmark is open- or closed-book** — for
   closed-book ones (like healthbench-professional) **every harness role you
   commission — planner, workers, AND reviewer — is denied
   WebSearch/WebFetch/Bash/browser** (the runtime hard-blocks retrieval for all
   three; only *you* keep web tools, for your own research). Their **Bash/shell
   is denied too** (a curl→web backdoor) — a commissioned role's trajectory
\end{Verbatim}
\end{minipage}
\end{lrbox}
\fcolorbox{black!30}{black!3}{\usebox{\setupSkillBox}}
\endgroup
\caption{Complete setup-agent skill (continued).}
\label{fig:setup-full-6}
\end{figure*}
\clearpage

\begin{figure*}[p]
\centering
\begingroup
\setlength{\fboxsep}{8pt}
\setlength{\fboxrule}{0.4pt}
\begin{lrbox}{\setupSkillBox}
\begin{minipage}{\dimexpr\textwidth-2\fboxsep-2\fboxrule\relax}
\raggedright
{\sffamily\bfseries Setup-agent process skill (continued)\par}
\smallskip
\begin{Verbatim}[fontsize=\normalsize,breaklines=true,breakanywhere=true,breaksymbolleft={},breaksymbolright={},breakautoindent=true]
   showing a Bash attempt return *"Bash exists but is not enabled in this
   context"* means **denied, not available**: don't read it as a working shell,
   and don't generalize your OWN (working) Bash to the roles you commission.
   For a legitimate local check (e.g. is an attachment present?) those roles
   use **Read/Glob/Grep**, which ARE available — never Bash. If closed-book,
   author every role for answering from the model's own knowledge — never
   instruct any of them to fetch URLs, verify sources online, or use web
   calculators, and don't judge their output as if they could. **Also mine the
   public benchmark guidelines/docs for how responses are scored at the
   protocol level** — general output/format conventions and any published
   penalties (e.g. an over-length / verbosity penalty, a required structure) —
   and if you find relevant **benchmark protocols**, **strengthen the skills
   with them** (public protocol, not the per-instance rubric). Do NOT go
   looking for the **per-instance** grading — the grader source, rubric text,
   or answer keys (sandboxed anyway); inferring "what done means" from the task
   itself is the point. From that, author a first `benchmark-planning`
   (**single-worker default**), `benchmark-worker`, and `benchmark-review` —
   general process, no dataset or grader/judge specifics. Initialize
   `scratchpad_setup.md` + `memory_setup.md`.
2. **Select** — `list_instances` → `describe_instance` on candidates → pick
   ~10 diverse, distinct instances (see Cost discipline). Reuse this SAME set
   to compare approaches fairly.
3. **Run the current approach** — `run_instances(ids)`; note the returned live
   trajectory paths.
4. **Observe** — while the run is in flight, use `wait` (as many times as
   needed) or tail the live trajectory; read `get_output` once instances
   finish, or early-cancel per Cost discipline.
5. **Judge + diagnose in parallel** — read each instance's output
   (`get_output`) and trajectory (`locate_trajectory` per role) and decide
   **yourself** whether harness did the task right — you do NOT get a grade.
   For the ones whose output looks wrong or weak, spawn subagents to work them
   concurrently and identify which **agent** got it wrong and the **general**
   weakness (planner over-decomposed or misread the task? worker skipped
   verifying its own work? reviewer approved without really checking / without
   writing its own tests?).
6. **Improve** — fold a general fix into the right skill and keep iterating
   **within this approach** — re-run, re-diagnose, refine — until you have real
   conviction in it. Only when the evidence says the strategy itself is wrong
   for this benchmark do you switch to a **different approach** (e.g. collapse
   a decomposition back to a single worker, or vice-versa): log the hypothesis
   in `scratchpad_setup.md`, change the skills, and test the new approach fully
   before judging it (see "one approach at a time — never in parallel"). Keep
   skills tight and non-contradictory; consolidate, don't pile on.
7. **Verify** — re-run the affected instances; confirm via trajectory + output
   that the behavior changed and the work looks better by your judgment,
   without degrading a case that previously looked right.
8. **Record** — update `scratchpad_setup.md` continuously; at the END of the
\end{Verbatim}
\end{minipage}
\end{lrbox}
\fcolorbox{black!30}{black!3}{\usebox{\setupSkillBox}}
\endgroup
\caption{Complete setup-agent skill (continued).}
\label{fig:setup-full-7}
\end{figure*}
\clearpage

\begin{figure*}[p]
\centering
\begingroup
\setlength{\fboxsep}{8pt}
\setlength{\fboxrule}{0.4pt}
\begin{lrbox}{\setupSkillBox}
\begin{minipage}{\dimexpr\textwidth-2\fboxsep-2\fboxrule\relax}
\raggedright
{\sffamily\bfseries Setup-agent process skill (continued)\par}
\smallskip
\begin{Verbatim}[fontsize=\normalsize,breaklines=true,breakanywhere=true,breaksymbolleft={},breaksymbolright={},breakautoindent=true]
   session, distill the durable lessons + the winning approach into
   `memory_setup.md`.
9. **Repeat** — new approaches / new failure areas until the budget is spent or
   improvement plateaus.

## Scratchpad and memory — two distinct files
- **`scratchpad_setup.md` — your LIVE experiment log.** Working notes you
  update continuously *this* session: the current hypothesis, the approach
  under test, per-instance observations, and the **run ledger** (every instance
  run + which approach it belonged to, so you respect the
  25/approach·100/session budget and never re-run one by accident). It is
  per-session (reset each run) and is **not** shipped — it's how you think, not
  the product.
- **`memory_setup.md` — the distilled record, written at the END of the
  session.** The durable takeaways a later `--improve` run starts from, kept
  **concise and sectioned**:
  - **Winning approach** — the strategy that worked best for this benchmark
    (single worker / decompose when X / …) and, briefly, what you ruled out
    and why.
  - **Lessons** — one entry per lesson/area: symptom → the role it belonged to
    → the general fix (and which skill/section) → status (fixed / open /
    hardened).

  `memory_setup.md` **is committed** alongside the skills (via `bench
  promote`), so — like the skills — keep it free of verbatim dataset/grader
  content: reference instance ids and describe general symptoms/fixes, but
  never paste rubric text, expected answers, or hidden test contents. Keep it
  tight — prune superseded notes rather than letting it grow.

## When you're done
Stop when the budget is exhausted or improvement plateaus — but a "plateau" in
**your own judgment on a narrow slice is not proof the skills are good**;
before you conclude, make sure you've stressed the diverse and hard/adversarial
cases, not just the ones that already look right. **In particular, never ship a
reviewer whose catch-and-fix path you have not watched fire.** A reviewer that
only ever approves good drafts is *untested* — you've validated the trivial
path, not the one that matters. Deliberately put a genuinely deficient draft in
front of it (stress a hard case, or hand it a weak worker output) and confirm
it **detects the specific defect and drives a fix** — a follow-up worker via
`submit_task_graph`, or an inline correction — then re-verify. If it
rubber-stamps the deficient draft, the reviewer skill is broken: strengthen it
until it catches. And if your outputs are all coming out "correct" and the
reviewer never needs to act, treat that as a sign you haven't stressed hard
enough — not that the skills are done. Leave the skills in a clean, shippable
state and `memory_setup.md` up to date. Report a short summary: what you
changed, what improved, and what remains open.
\end{Verbatim}
\end{minipage}
\end{lrbox}
\fcolorbox{black!30}{black!3}{\usebox{\setupSkillBox}}
\endgroup
\caption{Complete setup-agent skill (continued).}
\label{fig:setup-full-8}
\end{figure*}
\clearpage

\end{document}